\documentclass{article}
\usepackage{arxiv}

\usepackage[utf8]{inputenc} 
\usepackage[T1]{fontenc}    
\usepackage{hyperref}       
\usepackage{url}            
\usepackage{booktabs}       
\usepackage{amsfonts}       
\usepackage{nicefrac}       
\usepackage{microtype}      
\usepackage{lipsum}
\usepackage{graphicx}
\graphicspath{ {./images/} }
\usepackage[T1]{fontenc}

\usepackage{amsmath}
\usepackage{amssymb}
\usepackage{mathtools}
\usepackage{comment}
\usepackage{bm}

\usepackage{algorithm}
\usepackage{algpseudocodex}
\usepackage{accents}
\usepackage{placeins}

\usepackage[capitalize]{cleveref}
\crefformat{equation}{(#2#1#3)}
\Crefformat{equation}{Equation~(#2#1#3)}
\Crefname{equation}{Equation}{Eqs.}
\usepackage{subcaption}
\usepackage{booktabs}
\usepackage{multirow}

\usepackage{pifont}
\newcommand{\redx}{{\color{red}\ding{55}}}
\newcommand{\boldparagraph}[1]{\vspace{2pt}\noindent{\bf #1}}
\newcommand{\ie}{i.e.,\xspace}

\usepackage{xspace}

\usepackage{ulem}

\usepackage{etoolbox}
\apptocmd{\table}{\vspace{2.5pt}}{}{}
\pretocmd{\endtable}{\vspace{-4.5pt}}{}{}

\title{Compact Keyframe-Optimized Multi-Agent Gaussian Splatting SLAM}

\author{
Monica M.Q. Li \\
Department of Computer Engineering\\
Polytechnique Montreal\\
Montreal, QC, Canada \\
\texttt{monica.li@polymtl.ca} \\
\And
Pierre-Yves Lajoie \\
Department of Computer Engineering\\
Polytechnique Montreal\\
Montreal, QC, Canada \\
\texttt{pierre-yves.lajoie@polymtl.ca} \\
\And
Jialing Liu \\
Department of Mathematics\\
University of Wisconsin\\
Madison, WI 53706, USA \\
\texttt{jliu2535@wisc.edu} \\
\And
Giovanni Beltrame \\
Department of Computer Engineering\\
Polytechnique Montreal\\
Montreal, QC, Canada \\
\texttt{giovanni.beltrame@polymtl.ca} \\
}

\begin{document}
\maketitle
\begin{abstract}
Efficient multi-agent 3D mapping is essential for robotic teams operating in unknown environments, but dense representations hinder real-time exchange over constrained communication links. In multi-agent Simultaneous Localization and Mapping (SLAM), systems typically rely on a centralized server to merge and optimize the local maps produced by individual agents. However, sharing these large map representations, particularly those generated by recent methods such as Gaussian Splatting, becomes a bottleneck in real-world scenarios with limited bandwidth. We present an improved multi-agent RGB-D Gaussian Splatting SLAM framework that reduces communication load while preserving map fidelity. First, we incorporate a compaction step into our SLAM system to remove redundant 3D Gaussians, without degrading the rendering quality. Second, our approach performs centralized loop closure computation without initial guess, operating in two modes: a pure rendered-depth mode that requires no data beyond the 3D Gaussians, and a camera-depth mode that includes lightweight depth images for improved registration accuracy and additional Gaussian pruning. Evaluation on both synthetic and real-world datasets shows up to 85-95\% reduction in transmitted data compared to state-of-the-art approaches in both modes, bringing 3D Gaussian multi-agent SLAM closer to practical deployment in real-world scenarios. Code:  \url{https://github.com/lemonci/coko-slam}
\end{abstract}


\section{Introduction}
Simultaneous Localization and Mapping (SLAM) enables a device to build a map of an unknown environment while concurrently estimating its own pose. Visual SLAM (\ie SLAM using cameras) underpins robotic navigation and augmented reality (AR) and virtual reality (VR) applications. Recent advances in visual SLAM, both sparse and dense, have extended system capabilities in increasingly complex scenarios~\cite{mur2015orb, liao2022so, qin2018vins}. Sparse SLAM restricts itself to a handful of salient feature points, whereas dense SLAM reconstructs full-scene geometry with higher precision, with several applications like autonomous driving, drone navigation, and immersive virtual environments~\cite{huang2021di, bloesch2018codeslam, teed2021droid}. Traditional dense methods achieve robust tracking~\cite{whelan2012kintinuous, deng2023long} through explicit representations such as voxels, point clouds, or truncated signed distance functions (TSDF)~\cite{yao2019recurrent}, yet they struggle with capturing fine textures, filling scene gaps, and maintaining high-fidelity reconstructions.

The emergence of Neural Radiance Fields (NeRF) introduced an implicit representation that captures complex appearance and geometry with fine detail~\cite{yu2021pixelnerf, deng2022depth, guo2022nerfren, barron2021mip}. However, NeRF-based SLAM incurs heavy computational and memory demands, hindering real-time performance. To reconcile mapping accuracy with operational efficiency, 3D Gaussian Splatting (3DGS) has recently gained traction~\cite{kerbl20233d}. By modeling color, opacity, and orientation as continuous Gaussian distributions, 3DGS amalgamates explicit structural flexibility with implicit continuity, yielding faster rendering and improved geometric accuracy. Nonetheless, existing 3DGS SLAM implementations~\cite{yan2024gs, Matsuki:Murai:etal:CVPR2024, hu2024cg, keetha2024splatam} exhibit shortcomings in global optimization and loop closure detection, particularly over large-scale environments where accumulated drift undermines map consistency.

Multi-agent collaborative SLAM seeks to fuse the maps produced by independent agents into a globally consistent representation as they explore a shared environment~\cite{hu2023cp, yugay2024magicslammultiagentgaussianglobally}. Each agent constructs and optimizes its local map, while shared observations improve overall consistency and enable the generation of a cohesive global map that outperforms single-agent approaches in efficiency and geometric fidelity~\cite{schmuck2019ccm}. Despite these advantages, current multi-agent 3DGS SLAM approaches still lack the efficiency required for real-time collaborative mapping in large-scale environments. Specifically, multi-agent 3DGS SLAM systems often encounter the following challenges:

\boldparagraph{Keyframing:} 
Since retaining all sensor frames for mapping is computationally expensive, multi-agent visual SLAM systems must sample keyframes for tasks such as mapping, loop detection, and inter-robot pose estimation. Traditional heuristics—such as selecting every $n$-th frame~\cite{yugay2024magicslammultiagentgaussianglobally} or applying thresholds based on relative motion between frames~\cite{Matsuki:Murai:etal:CVPR2024}—often result in redundant keyframes while potentially discarding highly informative ones.

\boldparagraph{Submap size reduction:} 
To further improve efficiency, 3DGS SLAM systems group keyframes into \textit{submaps} that represent regions of the environment. However, the resulting 3D Gaussian models often include large, highly transparent Gaussians that occupy as much storage as more visually informative points while contributing little to the overall model quality. In multi-agent settings and large-scale environments, these low-information points often dominate the representation, leading to scalability challenges~\cite{zhang2024gaussianspa}. Without effective submap size reduction techniques, multi-agent 3DGS SLAM systems struggle to scale efficiently.

\begin{figure*}[ht!]
\vspace{0.5em}
    \centering
 \begin{subfigure}[b]{0.32\linewidth}
        \centering
        \includegraphics[width=\linewidth]{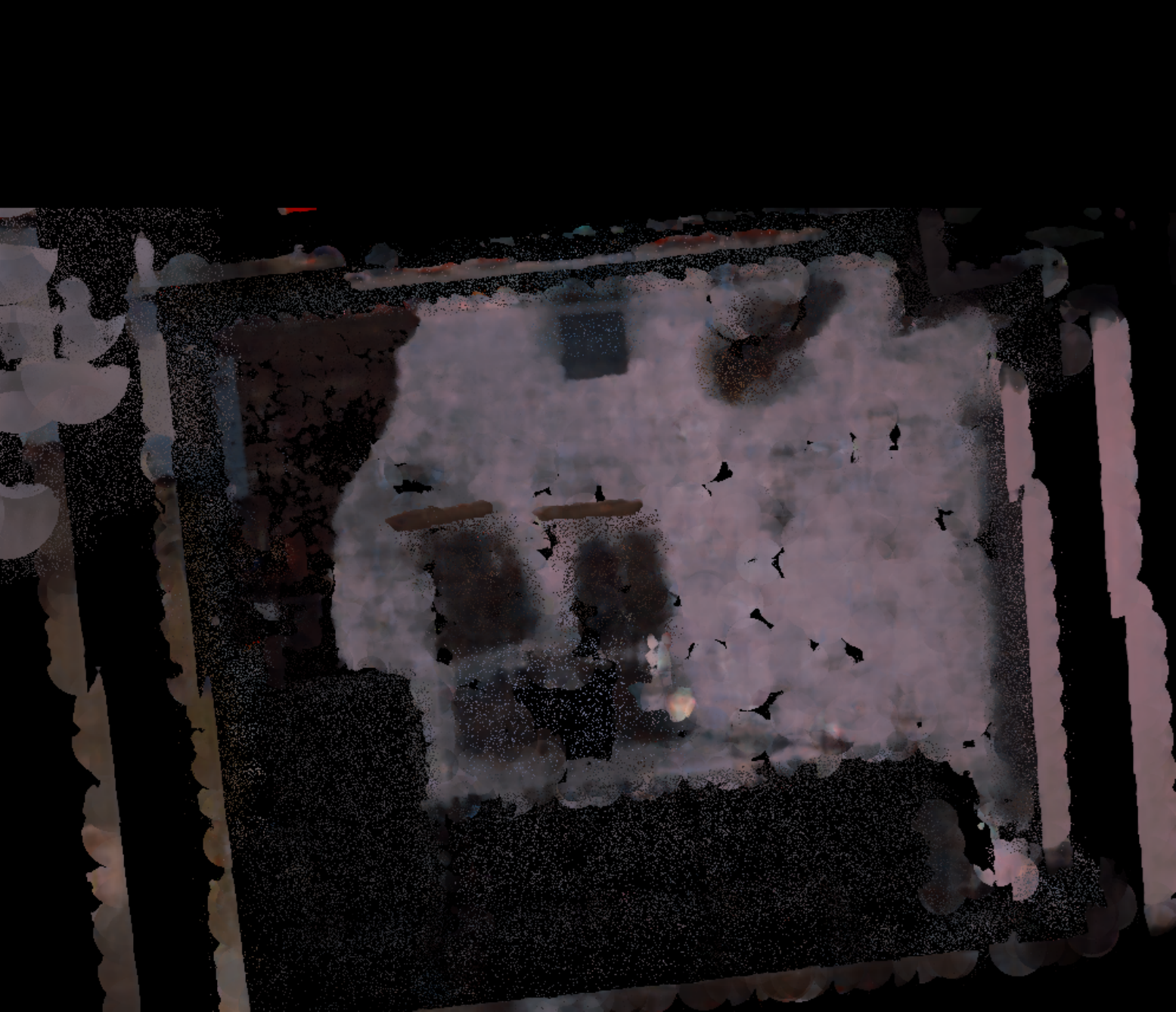}
        \caption{CP-SLAM}
    \end{subfigure}\hfill
    \begin{subfigure}[b]{0.32\linewidth}
        \centering
        \includegraphics[width=\linewidth]{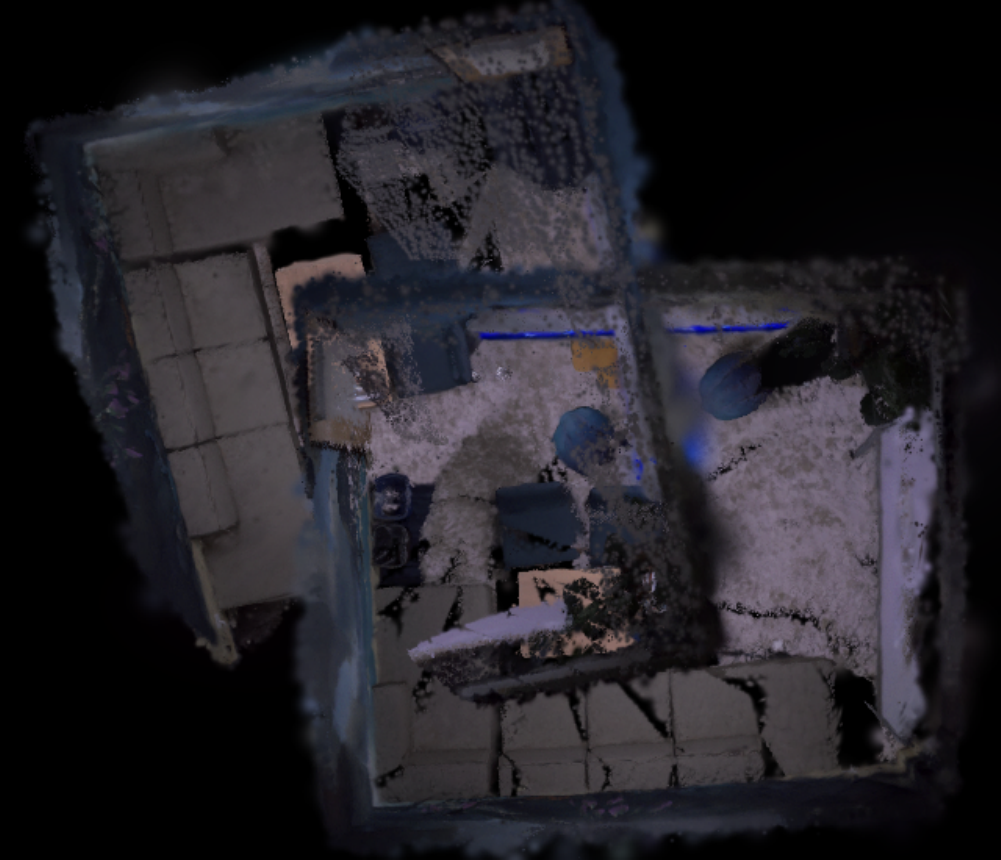}
        \caption{MAGiC-SLAM}
    \end{subfigure}\hfill
    \begin{subfigure}[b]{0.32\linewidth}
        \centering
        \includegraphics[width=\linewidth]{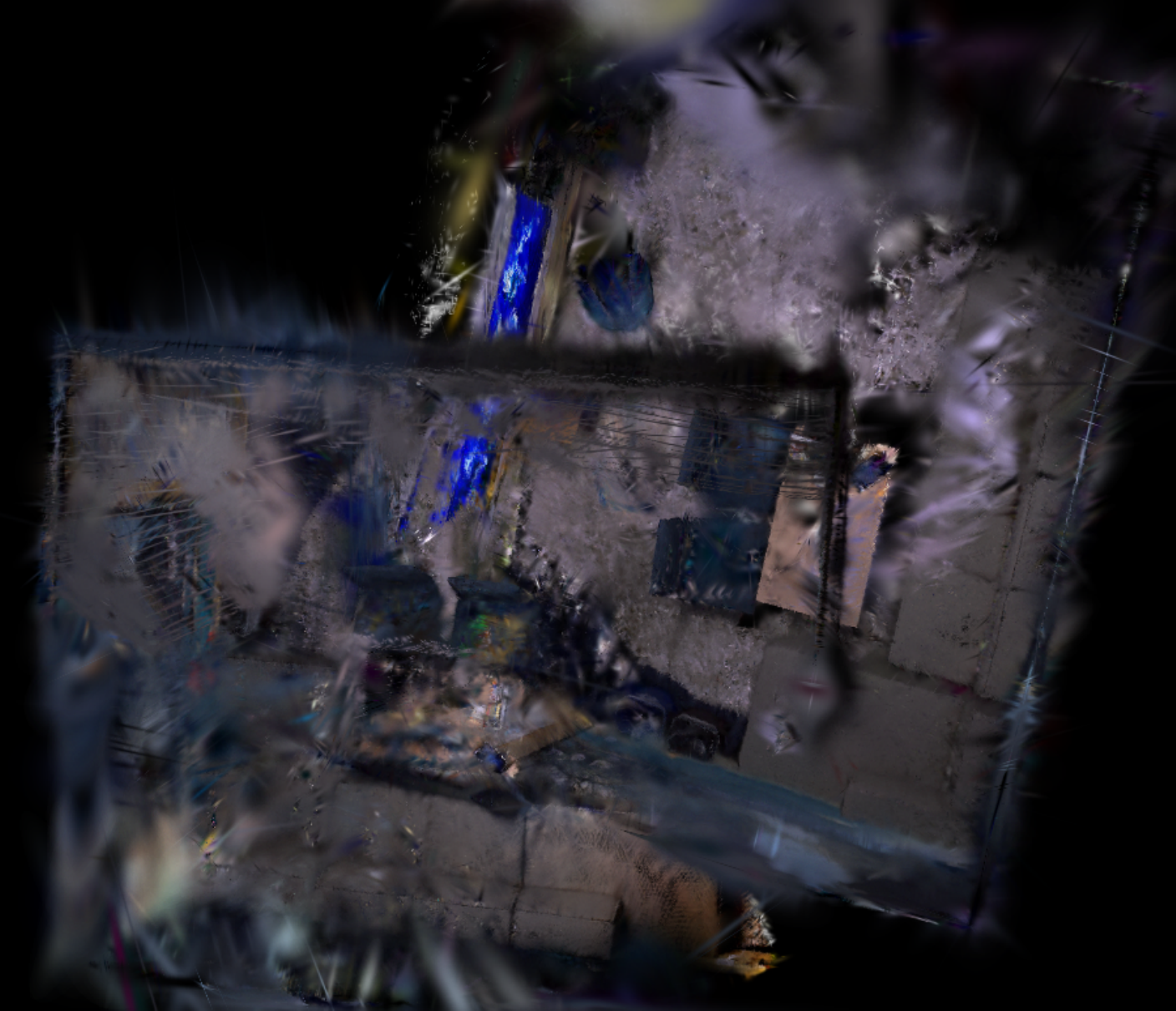}
        \caption{MAC-Ego3D}
    \end{subfigure}

    \begin{subfigure}[b]{0.32\linewidth}
        \centering
        \includegraphics[width=\linewidth]{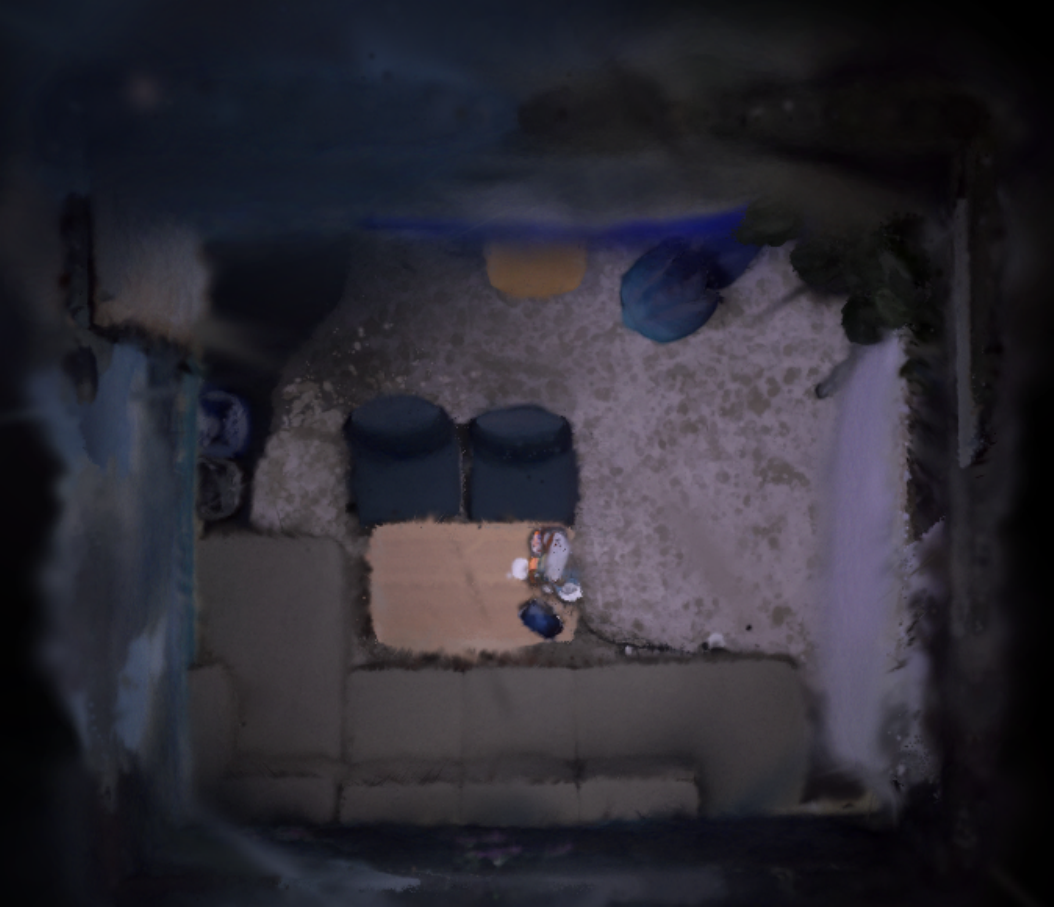}
        \caption{Ours (rendered depth)}
    \end{subfigure}\hfill
    \begin{subfigure}[b]{0.32\linewidth}
        \centering
        \includegraphics[width=\linewidth]{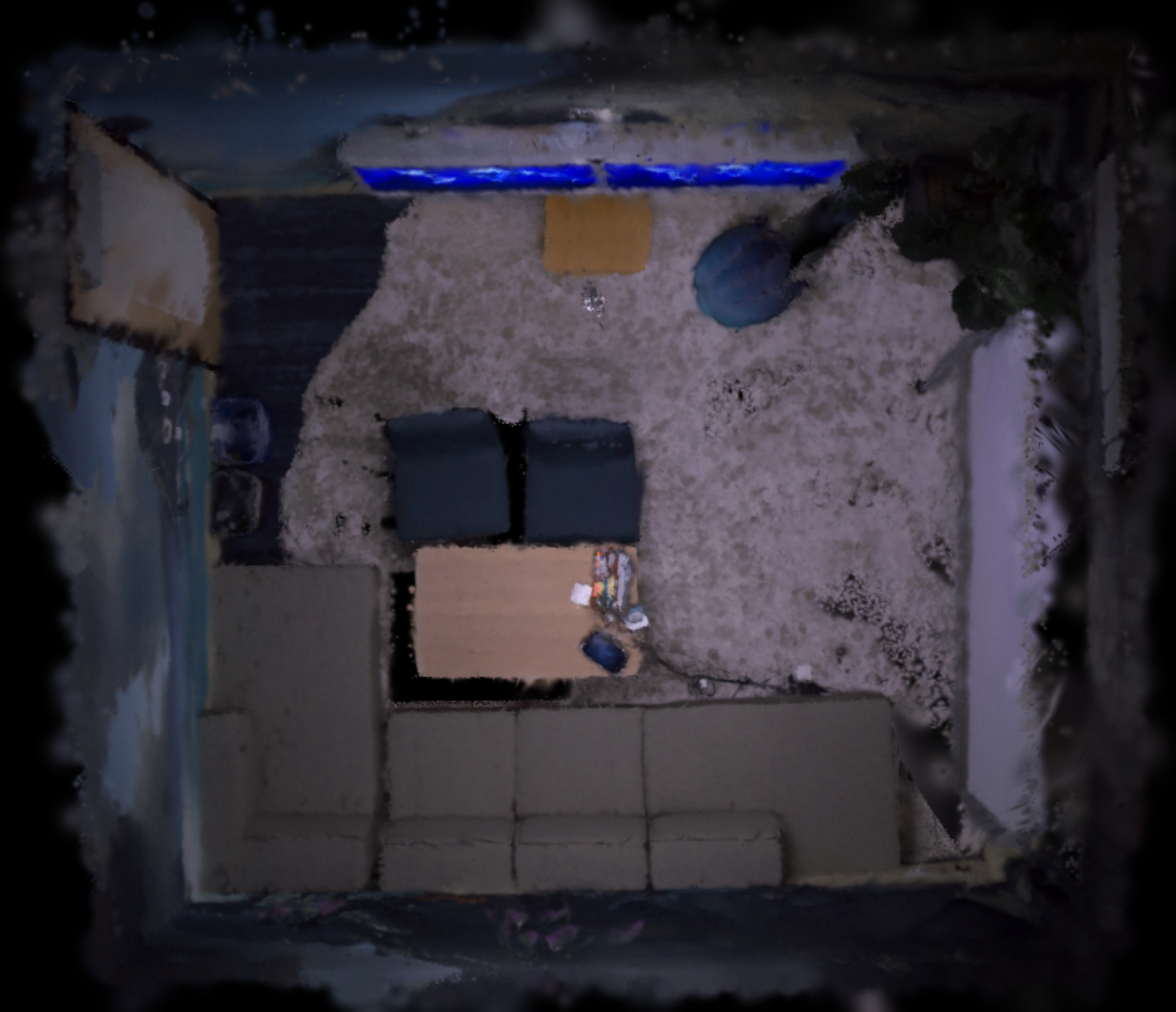}
        \caption{Ours (camera depth)}
    \end{subfigure}\hfill
    \begin{subfigure}[b]{0.32\linewidth}
        \centering
        \includegraphics[width=\linewidth]{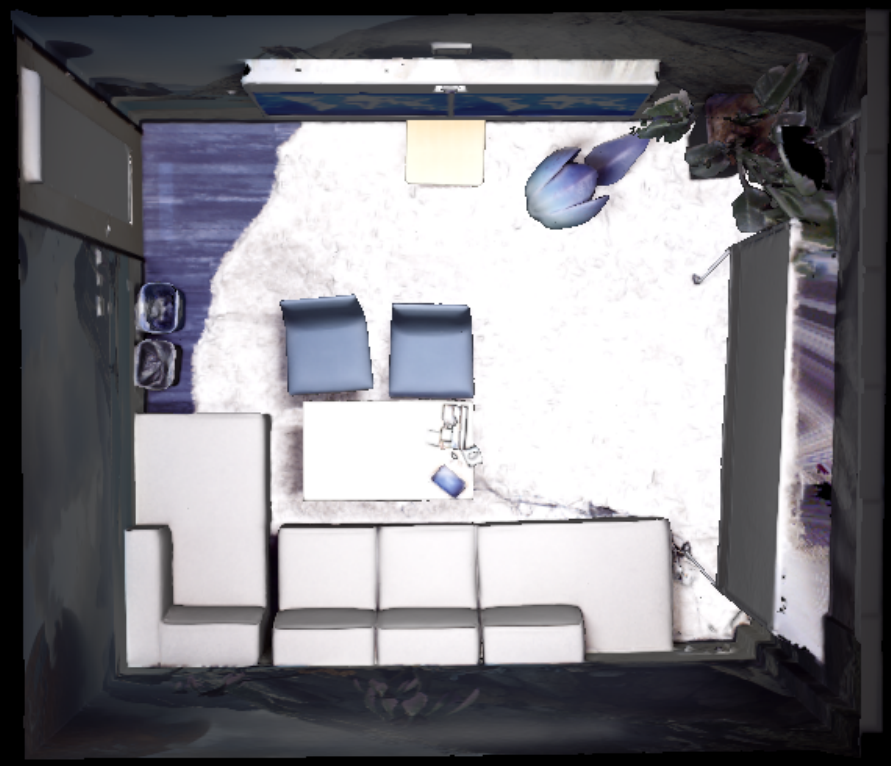}
        \caption{Ground truth}
    \end{subfigure}
    
    \caption{Top view comparison of merged maps of two agents in scene office-0 of the Replica dataset~\cite{hu2024cg}. Our method, Coko-SLAM, merges the submaps from different agents correctly while CP-SLAM~\cite{hu2023cp} (a), MAGiC-SLAM~\cite{yugay2024magicslammultiagentgaussianglobally} (b), and MAC-Ego3D~\cite{xu2025mac} (c) fail without initial relative poses between agents. The maps in (a) and (d) are merged using rendered images from the submaps; the maps in (b), (c), and (e) are merged using depth images directly from the camera.}
    \label{fig:topview_comparison}
    \vspace{-15pt}
\end{figure*}

\boldparagraph{Inter-robot loop closure and pose estimation:} Prior methods typically assume access to odometry or known initial poses, and estimate loop-edge constraints through refinement techniques such as ICP~\cite{yugay2024magicslammultiagentgaussianglobally, xu2025mac} or rendered loss optimization~\cite{xu2025mac, zhu2024loopsplat}. As a result, they fail to merge maps when such prior information is unavailable, as shown in \cref{fig:topview_comparison}. 

We propose Coko-SLAM, a compact keyframe optimized SLAM system on multi-agent Gaussian mapping, to reduce submap size and rely on pure 3D Gaussian submaps with keyframe feature vectors for loop closures. The agents only transmit keyframe feature vectors, camera poses, and 3D Gaussian submaps to the central server, without relying on point cloud nor images. Experimental results demonstrate that Coko-SLAM outperforms state-of-the-art approaches in rendering quality on the Replica and Aria multi-agent datasets, while reducing transmitted data and storage requirements by up to 95\%. 

The contributions of this work are summarized as follows:
\begin{enumerate}
    \item We implement an inter-robot loop closure system able to merge 3D Gaussian submaps even without camera images or initial relative poses between agents;
    \item We incorporate an optimization-sparsification strategy into 3DGS SLAM to eliminate redundant 3D Gaussians without degrading rendering quality;
    \item We adapt keyframe selection from LiDAR to visual SLAM, based on the marginal information provided by feature vectors.
\end{enumerate}

Coko-SLAM makes it feasible to deploy visual 3DGS multi-agent SLAM on robotic platforms with limited bandwidth. Since~\cite{lajoie2024collaborative} shows that most communication in multi-robot SLAM arises from data association, Coko-SLAM addresses this bottleneck through keyframe selection and submap compaction, allowing submap transmission to complete in several seconds rather than minutes, compared to MAGiC-SLAM~\cite{yugay2024magicslammultiagentgaussianglobally}.

\section{Background and Related Work}


\subsection{Multi-Agent SLAM}

Multi-agent SLAM extends the operational scale of SLAM via collaborative mapping, but it also introduces complexity and bandwidth constraints. Centralized frameworks such as CVI-SLAM~\cite{karrer2018cvi} and CCM-SLAM~\cite{schmuck2019ccm} fuse submaps on a server using pose-graph optimization but incur high communication overhead and single-point failure risk. Decentralized approaches like Swarm-SLAM~\cite{lajoie2023swarm} delegate loop closure and map fusion peer-to-peer, improving robustness but limited by onboard compute and inter-agent communication constraints. 

Regarding multi-agent SLAM with novel view synthesis(NVS), CP-SLAM~\cite{hu2023cp} applies neural mapping to multi-agent data but inherits high GPU costs and slow synthesis. MAC-Ego3D~\cite{xu2025mac} adapted fastGICP~\cite{koide2021voxelized} for real-time tracking and loop closure followed by photometric loss for refinement. This method requires the initial relative poses between agents for loop closure, and a large discrepancy between loop edges can lead to error because ICP requires the maps to be coarsely aligned initially. MAGiC-SLAM~\cite{yugay2024magicslammultiagentgaussianglobally} uses 3D Gaussians for submap merging and loop closure detection yet still requires agents to send point clouds to the server, maintaining heavy network load and under-utilizing local collaboration. 

We present a recap of MAGiC-SLAM~\cite{yugay2024magicslammultiagentgaussianglobally} as the base of our pipeline. During tracking, MAGiC-SLAM~\cite{yugay2024magicslammultiagentgaussianglobally} employs a hybrid implicit frame-to-frame and frame-to-model strategy. First, a coarse pose initialization uses deterministic dense registration via multi-scale ICP~\cite{besl1992method}: at each scale \(s\in\{1,\dots,S\}\), voxelized point clouds \(\mathcal{P}_{t},\mathcal{P}_{t-1}\) are aligned by minimizing the joint loss of rendered color and depth according to the camera RGB-D images. Next, pose refinement freezes Gaussian parameters and minimizes the re-rendering loss based on color and depth errors. In the refinement, pixels with low opacity or high residuals are masked out to focus on well-reconstructed regions. 

In the mapping process, each agent segments its RGB-D input into submaps, each represented as a collection of 3D Gaussians seeded at point-cloud samples and in low opacity regions. The camera poses are fixed and new Gaussians are optimized jointly based on~\cref{eq:joint_loss}.
\begin{equation}
    \mathcal{L} = \lambda_\text{color} \cdot \mathcal{L}_\text{color} + \lambda_\text{depth} \cdot \mathcal{L}_\text{depth} + \lambda_\text{reg} \cdot \mathcal{L}_\text{reg} \enspace,
\label{eq:joint_loss}
\end{equation}
where all $\lambda_{*} \in \mathbb{R}$ are hyperparameters.

The color loss $\mathcal{L}_\text{color}$ is defined as $\mathcal{L}_\text{color} = (1 - \rho) \cdot |\hat{I} - I|_1 + \rho \big(1 - \mathrm{SSIM}(\hat{I}, I) \big)$, where $\hat{I}, I$ are the rendered and input images respectively, and $\rho \in [0,1]$ is a blending weight. SSIM is the structural similarity index measure that assesses the perceived similarity of an image pair.

The depth loss $\mathcal{L}_\text{depth}$ is formulated as $\mathcal{L}_\text{depth} = |\hat{D} - D|_1$, where $\hat{D}, D$ are the rendered and input depth images.

The regularization loss $\mathcal{L}_\text{reg}$ is represented as $\mathcal{L}_\text{reg} = |K|^{-1} \sum_{k \in K}|s_k - \overline{s}_k|_1$, where $s_k\in\mathbb{R}^3$ is the scale of a 3D Gaussian, $\overline{s}_k$ is the mean submap scale, and $|K|$ is the number of Gaussians in the submap. This is a penalty term to prevent the elongation of Gaussians along the principal axis, as discussed in~\cite{Matsuki:Murai:etal:CVPR2024}.

After each agent finishes mapping, they send the submaps containing keyframe features, 3D Gaussians, odometry poses and the first keyframes' point cloud of the submaps to the server for loop closure and map merging.


\begin{figure*}[t!]
    \centering
    \includegraphics[width=\linewidth]{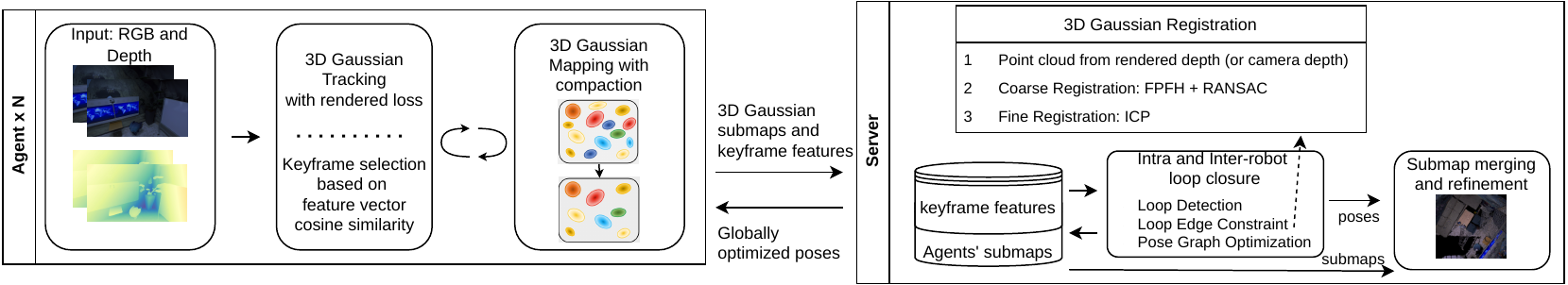}
    \caption{We adapt a keyframing method to select keyframes based on distance in feature space. We then eliminate redundant 3D Gaussians during mapping. The agents then send pure 3D Gaussian submaps with keyframe feature vectors to a server for loop closure and map merging, without the initial relative poses among agents.}
    \label{fig:architecture}
    \vspace{-15pt}
\end{figure*}

\subsection{Keyframing}

Keyframing is the decision process to save or discard the sensor input at a specific moment. It must balance accuracy with memory usage since keyframes are used as the source for localization, mapping and loop-closure detection but cannot be saved indefinitely~\cite{kretzschmar2011efficient}.

The keyframing methods in current 3DGS SLAM systems largely rely on heuristics such as Euclidean distance from other keyframes~\cite{Matsuki:Murai:etal:CVPR2024, yugay2023gaussianslam}, or simply
selecting every $n$-th frame as a keyframe~\cite{keetha2024splatam, yugay2024magicslammultiagentgaussianglobally}. These keyframing methods rely on environment-specific or expert tuning and can lead to sets of keyframes that are redundant or insufficient for localization.

Recently, research in LiDAR scan has provided new ideas for keyframing. Thorne et al.~\cite{thorne2024submodular} proposed keyframe selection and submap generation strategies for localization and map summarization. In their work, a keyframe is selected based on a distance threshold in feature space. Then the marginal value of each keyframe candidate is calculated and the selection provides guarantees on the quality of the chosen subset relative to the optimal selection. Though their approach has components specifically related to LiDAR, this keyframing strategy also provides valuable guidance for keyframing in visual SLAM.

\subsection{3D Gaussian compaction}

In a centralized system with multiple agents, each agent needs to transmit submaps to a server for map merging, via inter-robot loop closure and pose graph optimization (PGO).

In many realistic scenarios, the transmission from agents to server may fail due to distance, communication issues, or bandwidth limitations of the network. This failure will cause the server to work on incomplete data and can cause the map to be reconstructed incorrectly~\cite{mielle2019auto}, or even cause an agent to can get incorrect pose estimations from the server, which in turn can
affect the next step of exploration~\cite{yu2020review}. Therefore, reducing the size of submaps without sacrificing the quality on the agents' side before sending them to the server is meaningful as it saves bandwidth and can improve the reliability and efficiency of map merging.

Existing methods dedicated to storage reduction of 3D Gaussians can be classified into two categories: compression, which focuses on reducing data transmitted, and compaction, which aims to minimize the number of Gaussians. For more information on 3DGS compression and compaction algorithms, we recommend reading 3DGS.zip~\cite{bagdasarian20243dgs}.

In this work, we adopt a real-time compaction algorithm, GaussianSPA~\cite{zhang2024gaussianspa}, to remove redundant Gaussians with an optimization-sparsification method. GaussianSPA ranked the first place compaction method in the evaluation~\cite{bagdasarian20243dgs}.

We did not apply compression because most of these methods take several minutes to compress a single submap so they are not suitable for real-time SLAM tasks. The only currently available real-time compression method, FCGS~\cite{fcgs2025}, was trained on the original 3DGS work~\cite{kerbl20233d} and when applied to 3DGS scenes from other sources, reduces map quality.

\subsection{Loop closure for 3DGS SLAM}

Loop closures are essential to correct accumulated drift in SLAM. Some works on 3D Gaussian registration also provide guidance for loop closure and pose graph optimisation in 3D Gaussian maps. For example, registration-oriented methods such as GaussReg~\cite{Chang2024GaussReg} explore coarse-to-fine Gaussian splatting alignment but do not perform explicit loop closure. 

Most existing 3DGS SLAM loop-closure methods depend on known initial relative poses and odometry between agents~\cite{yugay2024magicslammultiagentgaussianglobally, xu2025mac, zhu2024loopsplat}. The estimation of the loop edge constraints in these approaches typically relies on fine-registration methods such as ICP~\cite{yugay2024magicslammultiagentgaussianglobally, xu2025mac} or rendered loss optimization~\cite{xu2025mac, zhu2024loopsplat}. Consequently, these methods cannot successfully merge maps when initial poses between agents are unknown, as demonstrated in \cref{fig:topview_comparison}. However, in real-world multi-agent SLAM scenarios, initial relative poses between agents are difficult to measure accurately and are typically unavailable in advance.

\section{Method}
\label{sec:methods}

The framework of Coko-SLAM is shown in \cref{fig:architecture}. Building on MAGiC-SLAM's codebase, we made three key improvements: we enhanced the keyframe selection method, adopted a compaction technique during 3D Gaussian training to reduce submap size, and implemented loop closure using pure 3D Gaussian submaps without requiring prior knowledge of the initial relative poses between agents.

\subsection{Keyframing using Feature Vector}
\label{subsec:keyframe_identification}

Selecting keyframes from the incoming data stream requires balancing two objectives: localization accuracy and memory efficiency. For localization, each candidate keyframe should have sufficient overlap with previously stored keyframes to ensure reliable tracking. In respect of memory usage, adding a new keyframes increases storage requirements. 

Adapted from the keyframe selection method for LiDAR scans~\cite{thorne2024submodular}, our approach relies on a pretrained neural network to efficiently determine the similarity between any two frames. The operator $\phi(\cdot)$ is the feature extractor to convert the frame into the feature space. In 3DGS SLAM, since the incoming frames are images on the edge-devices, we used Dino V2-Small~\cite{oquab2023dinov2} to get the feature vectors. 



Inspired by ~\cite{thorne2024submodular}, our online keyframing strategy operates as follows: given the set of previous keyframes in the submap $\mathcal{K}$, a current incoming frame $E$, and a keyframing threshold $\alpha$, we first compute the minimum feature distance $d = \min_{K \in \mathcal{K}} ||\phi(E)-\phi(K)||$ between the current frame's feature $\phi(E)$ and all existing keyframe features $\phi(K)$. If this minimum distance satisfies $d \geq \alpha$, the current frame is selected as a keyframe. Otherwise, it is not selected. 



As the number of keyframes decreases, we also need to adjust the number of submaps; otherwise, too few keyframes per submap will hinder the optimization of new Gaussians. In alignment with the original MAGiC-SLAM configuration, we set the number of keyframes per submap to 10.

\subsection{3D Gaussian Compaction}

According to~\cite{zhang2024gaussianspa}, the 3DGS compaction can be treated as an optimization problem with respect to the training loss $\mathcal{L}$ with a constraint on the number of Gaussians $\kappa$. As the constraint is not differentiable, this original problem cannot be directly optimized by a gradient-based method. However, it can be converted into an equivalent differentiable problem represented by~\cref{eq:obj-lagrangian} via dual Lagrangian multiplier~\cite{zhang2024gaussianspa}:
\begin{equation}\label{eq:obj-lagrangian}
  L(\bm{a},\bm{z},\bm{\Theta},\bm{\lambda};\delta)=\mathcal{L}(\bm{a},\bm{\Theta})+h(\bm{z})+\frac{\delta}{2}\|\bm{a}-\bm{z}+\bm{\lambda}\|^2+\frac{\delta}{2}\|\bm{\lambda}\|^2,
\end{equation}
where $\bm{a}$ represents the gaussian opacity and $\bm{\Theta}$ represents the other 3DGS variables, $\bm{\lambda}$ is the dual Lagrangian multiplier, and $\delta$ is the penalty parameter. Correspondingly, we can separately optimize over $\bm{\Theta}$ and $\bm{a}$, with the auxiliary variable $\bm{z}$ in the augmented Lagrangian. Then two split sub-problems can be solved in alternately optimization and sparsification steps, as in \cref{alg:optimization-sparsification}.
\label{subsec: 3dgs_compaction}

\begin{algorithm}[ht!]
\newcommand{\assign}{$\leftarrow$}
\caption{Procedure of Optimization-Sparsification}
\label{alg:optimization-sparsification}
\begin{algorithmic}[1]
\renewcommand{\algorithmicrequire}{\textbf{Input:}}
\renewcommand{\algorithmicensure}{\textbf{Output:}}  
    \Require Gaussian opacity $\bm{a}$, other 3D Gaussian variables $\bm{\Theta}$, target number of Gaussians $\kappa$, penalty parameter $\delta$, feasibility tolerance $\epsilon$, 3DGS training loss $\mathcal{L}$, learning rate $\eta$, proximal operator $\mathrm{\bf{prox}}_h$, maximal iteration $T$, current iteration $t$. \par
    \Ensure Optimized $\bm{a}$ and $\bm{\Theta}$
    \State $\bm{z}$ \assign $\bm{a}$, $\bm{\lambda}$ \assign $\bm{0}$; $t$ \assign $0$;
    \While{$\|\bm{a} - \bm{z}\|^2 > \epsilon$ and $t \leq T$}
        \State Update $\bm{a}$ and $\bm{\Theta}$ with $ \\\bm{a}\leftarrow\bm{a}-\eta \frac{\partial L}{\partial \bm{a}},  ~~\bm{\Theta}\leftarrow\bm{\Theta}-\eta \frac{\partial L}{\partial \bm{\Theta}}$;
        \Comment{{\textit{Optimization}}}
        \State Update $\bm{z}$ with \\ $\bm{z}\leftarrow\mathrm{\bf{prox}}_h(\bm{a}+\bm{\lambda})$;
        \Comment{{\textit{Sparsification}}}
        \State Update $\bm{\lambda}$ with $ \\\bm{\lambda}\leftarrow\bm{\lambda}+\bm{a}-\bm{z}$;
        \Comment{{\textit{Multiplier Update}}}
        \State $t$ \assign $t+1$;
    \EndWhile
\end{algorithmic}
\end{algorithm}

The optimization, sparsification and multiplier update steps are performed alternately in the 3DGS training process until convergence in \cref{alg:optimization-sparsification}. Generally, the solution convergences once the condition, $\|\bm{a}-\bm{z}\|^2\le \epsilon$, is satisfied, or the iterations reach the predefined maximum number. We set $\epsilon=0.005$ following~\cite{zhang2024gaussianspa} and $T=950$ to be nearly the end of the training iterations $1000$.


While those steps are performed alternately, the process eventually converges to a sparse set that maximizes performance, as proved in~\cite{zhang2024gaussianspa}. Upon the convergence, some Gaussians' variables become zeros and the information can be preserved and transferred to the ``non-zero'' Gaussians. 

As the original GaussianSPA was run between the 15000 - 25200th iteration of the 30000 iterations in the original 3DGS training setup, it does not fulfill the real-time requirement of a SLAM system. As the sparsification needs to start at a later stage of the 3D Gaussian training when the gradients of the Gaussians do not change significantly, we start the optimization-sparsification at the 700th iteration and remove the zero Gaussians at the 950th iteration for real-time mapping, as the total training iterations are 1000.

\subsection{Loop closure between 3D Gaussian submaps}
\label{subsec: loop_closure}
Loop closure enforces global consistency by correcting past submap and keyframe poses when a revisit is detected. The process begins each time a new submap is finalized and proceeds through three stages: loop detection, loop edge constraint and pose graph optimization.

\boldparagraph{Loop detection:} In \cref{subsec:keyframe_identification}, we have already got the feature vectors of keyframes in each submap. For each keyframe $k$ in submap \(\mathcal{S}\), we first calculate the cosine similarity score of all keyframes within the same submap \(\mathcal{S}\), and rank them based on the similarity score, \ie for each keyframe $k$ we have a sorted list of all keyframe ids in the same submap. Then for each keyframe, we select the similarity of the $p-$th percentile keyframe in the sorted list as a threshold $s_{\text{self}}$ for keyframe $k$.

Then we can also calculate cosine similarities between keyframes of submap \(\mathcal{S}\) and keyframes of another submap \(\mathcal{T}\). We also choose the $p$-th percentile similarity as $s_{\text{cross}}$. If $s_{\text{cross}} > s_{\text{self}}$, a loop is identified.


\boldparagraph{Loop edge constraint:} The loop edge constraint can be framed as the alignment between two overlapping submaps, source submap \(\mathcal{S}\) and target submap \(\mathcal{T}\), as the estimation of a rigid transform \(T\) that brings \(\mathcal{S}\) into the coordinate frame of \(\mathcal{T}\). As the correspondences between Gaussians in different submaps are many-to-many instead of one-to-one, directly registering Gaussians treating their centers as points in point cloud is unreliable, as shown in~\cite{yugay2024magicslammultiagentgaussianglobally} and~\cite{zhu2024loopsplat}.

To perform intra and inter-robot registration in pure 3D Gaussian submaps, we first render the depth images $\hat{D}_\mathcal{S}$ and $\hat{D}_\mathcal{T}$ from the first keyframe views in \(\mathcal{S}\) and \(\mathcal{T}\). $\hat{D}_\mathcal{S}$ and $\hat{D}_\mathcal{T}$ can then be converted into point clouds $\hat{P}_\mathcal{S}$ and $\hat{P}_\mathcal{T}$. Following~\cite{rusu2009fast}, $\hat{P}_\mathcal{S}$ and $\hat{P}_\mathcal{T}$ are first downsampled to extract Fast Point Feature Histograms (FPFH) to present local geometric features in point cloud. A coarse transformation can be found by correspondence search via RANSAC based on FPFHs. The coarse transformation can then be refined by ICP~\cite{icp} onto the full-resolution point clouds.

We also provide results using the depth images $D_\mathcal{S}$ and $D_\mathcal{T}$ directly from the camera. The same registration methods are implemented onto point clouds $P_\mathcal{S}$ and $P_\mathcal{T}$ based on $D_\mathcal{S}$ and $D_\mathcal{T}$. As the depth is saved in the submap file and rendering is not needed for registration, similar to MAGiC-SLAM, the Gaussians in a submap can be removed if they are visible from the first keyframe view in the next submap, potentially reducing the sizes of submaps in certain scenes.

\boldparagraph{Pose graph optimization (PGO)}
We first construct a pose graph whose nodes \(\Delta \mathbf{X}_k\) represent relative corrections for each submap \(k\). Odometry edges connect consecutive submaps, initialized as identity constraints. Loop edges connect loop‑detected pairs \((k,j)\), with relative transform \(\mathbf{T}_{kj}\) and information matrix \(\mathbf{\Lambda}_{kj}\) derived from the registrations. The PGO optimizer then solve for all \(\{\Delta \mathbf{X}_k\}\) by minimizing 

\begin{align}
    \mathcal{L}_{\mathrm{PGO}} &=
    \sum_{\substack{(i,i+1)}} \bigl\|\log(\Delta \mathbf{X}_{i+1}^{-1}\,\Delta \mathbf{X}_i)\bigr\|^2_{\mathbf{\Lambda}_{i,i+1}} \nonumber \\
    &\quad + \sum_{(k,j)\in\mathcal{L}}
    \bigl\|\log(\mathbf{T}_{kj}^{-1}\,\Delta \mathbf{X}_k^{-1}\,\Delta \mathbf{X}_j)\bigr\|^2_{\mathbf{\Lambda}_{kj}}
\end{align}
using a robust line‑process formulation to downweight inconsistent edges via GTSAM~\cite{gtsam}.


In the server level, the PGO provides the rigid transformations between submaps from different agents. After all agents finish, the server performs a two-stage fusion of submaps into a unified global Gaussian map. In the coarse merging, cached submaps are concatenated, excluding Gaussians with zero opacity in the initial keyframe. Then during refinement, a brief optimization of Gaussian color and position via rendering losses removes visual and geometric artifacts, followed by pruning of zero-opacity Gaussians.

\section{Evaluation}
\boldparagraph{Datasets}
In alignment with MAGiC-SLAM~\cite{yugay2024magicslammultiagentgaussianglobally}, we test Coko-SLAM on the multi-agent Replica~\cite{hu2024cg} dataset, which contains four two-agent RGB-D sequences in a synthetic environment. Each sequence consists of 2500 frames, except for the Office-0 scene, which has 1500 frames.

Replica is a synthetic dataset without a test set for NVS. So we extend the evaluation to real-world scenes using the ego-centric Aria~\cite{pan2023aria} dataset, which provides ground-truth depth and camera information from recordings of two rooms in a real-world environment. Each Aria scene contains three agent sequences of 500 frames.

To eliminate the use of initial relative poses among agents, the trajectories for individual agents were converted to the local camera frame of the agent's first pose in the trajectory.

\boldparagraph{Baselines}
We evaluate rendering, keyframe counts and data transmitted with CP-SLAM~\cite{hu2023cp}, MAGiC-SLAM~\cite{yugay2024magicslammultiagentgaussianglobally} and MAC-Ego3D~\cite{xu2025mac} as they are state-of-the-art multi-agent NVS-capable SLAM systems. CP-SLAM used rendered RGB-D for loop closure, while MAC-Ego3D and MAGiC-SLAM used point cloud from camera depth images for loop-closure.
 All experiments were run on a server with a NVIDIA H100 GPU with 80GB VRAM.

\boldparagraph{Evaluation Metrics}
We assess map quality by rendering metrics: PSNR, SSIM~\cite{wangzhou2004image}, and LPIPS~\cite{zhang2018unreasonable}. For the evaluation of the robustness of rendering quality, we run experiments on 30 different seeds and report the mean with standard deviation in parentheses in \cref{tab:rendering_results} and \cref{tab:nvs_results}.

We also report the number of keyframes per agent in~\cref{tab:keyframe_count}, the total data transmitted, and the submap sizes in~\cref{tab:submap_sizes}. These are reported without standard deviation because the random seeds do not influence these results. As our tracking is the same as that of MAGiC-SLAM, tracking error is not reported.

\subsection{Results}

Our evaluation focuses on training view synthesis quality, novel view synthesis performance, and communication efficiency metrics. The best metrics are bolded and the second best metrics are underlined in the tables.

\boldparagraph{Training View Synthesis}~\cref{tab:rendering_results} presents the training view synthesis results on the Replica multiagent dataset. Our method achieves the best results on all Replica training view metrics. Our approach with camera depth achieves the highest PSNR and SSIM, outperforming all baselines by a substantial margin, and our rendered depth approach also achieves satisfactory map quality. In comparison, CP-SLAM shows moderate performance but struggles with depth accuracy, particularly in the Apart-1 scenario. MAGiC-SLAM fails completely on apartment scenarios, as indicated by the red crosses, demonstrating poor robustness to challenging environments. MAC-Ego3D, while more robust than MAGiC-SLAM, still underperforms our method across all metrics.

\begin{table}
    \caption{\textbf{Training view synthesis performance on Replica Multiagent dataset without initial relevant poses between agents.} The merged map based on multiple agents' submaps is evaluated by synthesizing training views. \redx{} indicates failures during map merging. \textbf{Bold} = best, \underline{underlined} = second best.}
    \centering
    \setlength\tabcolsep{4pt}
    \resizebox{\columnwidth}{!}{%
    \begin{tabular}{llrrrr}
    \toprule
    \textbf{Methods} & \textbf{Metrics} & \textbf{Office-0} & \textbf{Apart-0} & \textbf{Apart-1} & \textbf{Apart-2} \\
    \midrule
    \multirow{4}{*}{\textbf{CP-SLAM}}
    & PSNR [dB] $\uparrow$ & 24.257 (2.224) & 24.56 (2.502) & 19.185 (3.800) 	& 25.425 (4.209)  \\
    & SSIM $\uparrow$ & 0.749 (0.104) & 0.765 (0.099) & 0.646 (0.177) 	& 0.82 (0.173)\\
    & LPIPS $\downarrow$ & 0.434 (0.130) 	&0.437 (0.139) 	&0.555 (0.176) 	&0.403 (0.194)\\
    & Depth L1 [cm] $\downarrow$ &12.637 (12.727) 	&14.618 (14.746) 	&26.619 (27.384) 	&5.061 (5.222) \\
    \midrule
    \multirow{4}{*}{\textbf{MAGiC-SLAM}}
    & PSNR [dB] $\uparrow$ & 25.407 (0.237) & \redx & \redx & \redx \\
    & SSIM $\uparrow$ & 0.295 (0.005) & \redx & \redx & \redx \\
    & LPIPS $\downarrow$ & 0.839 (0.003) & \redx & \redx & \redx \\
    & Depth L1 [cm] $\downarrow$ & 20.536 (1.001) & \redx & \redx & \redx \\
    \midrule
    \multirow{4}{*}{\textbf{MAC-Ego3D}}
    & PSNR [dB] $\uparrow$ & 27.326 (0.924) & 30.431 (0.786) & 22.949 (3.169) & 27.983 (1.420) \\
    & SSIM $\uparrow$ & 0.912 (0.014) & 0.937 (0.006) & 0.853 (0.028) & 0.914 (0.017) \\
    & LPIPS $\downarrow$ & 0.201 (0.037) & 0.191 (0.014) & 0.282 (0.051) & \underline{0.200} (0.025) \\
    & Depth L1 [cm] $\downarrow$ & 23.563 (3.988) & 8.873 (2.170) & 26.928 (10.049) & 18.177 (2.711) \\
    \midrule
    \multirow{4}{*}{\textbf{Ours(Rendered depth)}}
    & PSNR [dB] $\uparrow$ & \underline{36.132} (2.837) & \underline{35.081} (0.337) & \underline{26.663} (0.902) & \underline{29.696} (0.330) \\
    & SSIM $\uparrow$ & \underline{0.975} (0.025) & \underline{0.967} (0.003) & \underline{0.906} (0.020) & \underline{0.942} (0.005) \\
    & LPIPS $\downarrow$ & \underline{0.083} (0.042) & \underline{0.132} (0.005) & \underline{0.248} (0.021) & 0.205 (0.007) \\
    & Depth L1 [cm] $\downarrow$ & \underline{0.753} (0.419) & \underline{1.243} (0.097) & \underline{7.823} (1.600) & \underline{2.398} (0.269) \\
    \midrule
    \multirow{4}{*}{\textbf{Ours(Camera Depth)}}
    & PSNR [dB] $\uparrow$ & \textbf{39.287} (0.295) & \textbf{36.634} (0.333) & \textbf{29.189} (0.297) & \textbf{31.072} (0.298) \\
    & SSIM $\uparrow$ & \textbf{0.991} (0.001) & \textbf{0.977} (0.002) & \textbf{0.945} (0.003) & \textbf{0.959} (0.003) \\
    & LPIPS $\downarrow$ & \textbf{0.059} (0.002) & \textbf{0.112} (0.003) & \textbf{0.201} (0.005) & \textbf{0.185} (0.005) \\
    & Depth L1 [cm] $\downarrow$ & \textbf{0.423} (0.018) & \textbf{0.829} (0.065) & \textbf{2.685} (0.290) & \textbf{1.474} (0.176) \\
    \bottomrule
    \end{tabular}%
    }
    \label{tab:rendering_results}
\end{table}

\boldparagraph{Novel View Synthesis}~\cref{tab:nvs_results} 
evaluates novel view synthesis capabilities on the Aria multiagent dataset. Our camera depth variant achieves the best performance on most metrics; the rendered depth variant leads on Novel View Room0 SSIM, and MAGiC-SLAM achieves the best Novel View Room1 SSIM. We notice that on the Aria dataset MAC-Ego3D had training view metrics substantially better than novel view performance. This performance gap indicates overfitting potentially caused by too few keyframes selected. In contrast, our method maintains consistent performance between training and novel views, with better generalization.

Our rendered depth results show relatively lower performance on Aria compared to Replica datasets. This is primarily attributed to the smaller image resolution in Aria ($512 \times 512$) compared to Replica ($1200 \times 680$). The reduced resolution amplifies the impact of noise in rendered depth, which affects registration accuracy and system performance.

\begin{table}
    \caption{\textbf{Novel view synthesis synthesis performance on Aria Multiagent dataset without initial relevant poses between agents.} The merged map based on multiple agents' submaps is evaluated by synthesizing novel views. The merged map based on multiple agents' submaps is evaluated by synthesizing novel and training views. \textbf{Bold} = best, \underline{underlined} = second best.}
    \centering
    \setlength\tabcolsep{4pt}
    \resizebox{\columnwidth}{!}{%
    \begin{tabular}{llrrrr}
    \toprule
    \multirow{2}{*}{\textbf{Methods}} & \multirow{2}{*}{\textbf{Metrics}} & \multicolumn{2}{c}{\textbf{Novel Views}} & \multicolumn{2}{c}{\textbf{Training Views}} \\
    \cmidrule(lr){3-4} \cmidrule(lr){5-6}
    & & \textbf{Room0} & \textbf{Room1} & \textbf{Room0} & \textbf{Room1} \\
    \midrule
    \multirow{4}{*}{\textbf{CP-SLAM}}
    & PSNR [dB] $\uparrow$ & 8.572 (0.286)  & 9.571 (0.368)  & 9.384 (0.632)  & 9.063 (0.814)  \\
    & SSIM $\uparrow$ & 0.292 (0.02)  & 0.268 (0.033)  & 0.94 (0.044)  & 0.959 (0.038)  \\
    & LPIPS $\downarrow$ & 0.879 (0.01)  & 0.914 (0.028)  & 0.208 (0.029)  & 0.257 (0.046)  \\
    & Depth L1 [cm] $\downarrow$ & 177.519 (10.136)  & 123.016 (7.009)  & 198.469 (197.717)  & 138.848 (138.472) \\
    \midrule
    \multirow{4}{*}{\textbf{MAGiC-SLAM}}
    & PSNR [dB] $\uparrow$ & 13.506 (0.125) & 19.092 (0.332) & 20.24 (0.074) & 22.772 (0.107) \\
    & SSIM $\uparrow$ & 0.531 (0.006) &\textbf{0.733} (0.016) & 0.759 (0.004) & 0.829 (0.004) \\
    & LPIPS $\downarrow$ & 0.627 (0.006) & \underline{0.392} (0.016) & 0.370 (0.006) & \underline{0.304} (0.007) \\
    & Depth L1 [cm] $\downarrow$ & 101.829 (3.370) & \underline{17.008} (0.947) & 27.053 (0.582) & \underline{14.217} (0.142) \\
    \midrule
    \multirow{4}{*}{\textbf{MAC-Ego3D}}
    & PSNR [dB] $\uparrow$ & 12.171 (0.233) & 17.500 (0.038) & \underline{23.030} (1.7039) & \underline{25.299} (1.867) \\
    & SSIM $\uparrow$ & 0.496 (0.010) & 0.573 (0.029) & 0.812 (0.016) & \underline{0.877} (0.019) \\
    & LPIPS $\downarrow$ & 0.713 (0.005) & 0.559 (0.035) & \underline{0.352} (0.054) & 0.310 (0.062) \\
    & Depth L1 [cm] $\downarrow$ & 94.709 (33.702) & 22.688 (2.462) & 41.363 (7.804) & 15.896 (2.968) \\
    \midrule
    \multirow{4}{*}{\textbf{Ours(Rendered depth)}}
    & PSNR [dB] $\uparrow$ & \underline{15.831} (1.536) & \underline{19.133} (1.243) & 22.143 (0.873) & 23.598 (2.326)  \\
    & SSIM $\uparrow$ &\textbf{0.668} (0.100) & 0.642 (0.068) & \underline{0.758} (0.035) & 0.818 (0.088)  \\
    & LPIPS $\downarrow$ & \underline{0.508} (0.098) & 0.437 (0.086) & 0.437 (0.040) & 0.343 (0.106)  \\
    & Depth L1 [cm] $\downarrow$ & \underline{54.406} (38.299) & 18.999 (21.477) & \underline{24.975} (2.272) & 14.330 (26.735) \\
    \midrule
    \multirow{4}{*}{\textbf{Ours(Camera Depth)}}
    & PSNR [dB] $\uparrow$ & \textbf{19.080} (0.995) & \textbf{20.527} (0.116) & \textbf{24.176} (0.592) & \textbf{25.892} (0.115) \\
    & SSIM $\uparrow$ & \underline{0.625} (0.047) & \underline{0.727} (0.007) & \textbf{0.889} (0.018) & \textbf{0.901} (0.003) \\
    & LPIPS $\downarrow$ & \textbf{0.480} (0.052) & \textbf{0.311} (0.007) & \textbf{0.262} (0.026) & \textbf{0.217} (0.008) \\
    & Depth L1 [cm] $\downarrow$ & \textbf{28.319} (9.257) & \textbf{8.696} (0.238) & \textbf{10.397} (5.632) & \textbf{3.034} (0.096) \\
    \bottomrule
    \end{tabular}%
    }
    \label{tab:nvs_results}
\end{table}

\boldparagraph{Communication Efficiency}~\cref{tab:keyframe_count} and \cref{tab:submap_sizes} analyze the communication efficiency through keyframe counts and submap sizes. Our method achieves competitive keyframe efficiency while maintaining rendering quality. On Replica datasets, our keyframe counts are comparable to or better than those of the baselines, while on Aria datasets, our counts are slightly higher than the most efficient baselines but still reasonable.

Regarding submap sizes~\cref{tab:submap_sizes}, our camera depth variant demonstrates an important trade-off characteristic. Camera depth submaps can be more compact than pure 3D Gaussian submaps (our rendered depth variant) because depth images from cameras are saved directly in submap files, eliminating the need for rendering during registration. This allows Gaussians visible from the first keyframe view in subsequent submaps to be removed, reducing storage requirements. However, this efficiency depends on scenario-specific co-visibility patterns: when co-visibility is low, fewer Gaussians will be removed, potentially making camera depth submaps larger than pure 3D Gaussian submaps.

Despite CP-SLAM and MAC-Ego3D achieving smaller submap sizes and keyframe counts on the Aria dataset, their rendering quality is unsatisfactory, highlighting the importance of balancing efficiency with performance quality.

\begin{table}
    \caption{\textbf{Number of keyframes per agent across different scenes.} Comparison for Replica and Aria datasets. - indicates that CP-SLAM does not support more than two agents. \textbf{Bold} = best (fewest), \underline{underlined} = second best.}
    \centering
    \setlength\tabcolsep{4pt}
    \resizebox{\columnwidth}{!}{%
    \begin{tabular}{lccccccccccccccc}
    \toprule
    & \multicolumn{8}{c}{\textbf{Replica}} & \multicolumn{6}{c}{\textbf{Aria}} \\
    \cmidrule(lr){2-9} \cmidrule(lr){10-15}
    & \multicolumn{2}{c}{\textbf{Office-0}} & \multicolumn{2}{c}{\textbf{Apart-0}} & \multicolumn{2}{c}{\textbf{Apart-1}} & \multicolumn{2}{c}{\textbf{Apart-2}} & \multicolumn{3}{c}{\textbf{Room0}} & \multicolumn{3}{c}{\textbf{Room1}} \\
    \textbf{Methods} & \textbf{A0} & \textbf{A1} & \textbf{A0} & \textbf{A1} & \textbf{A0} & \textbf{A1} & \textbf{A0} & \textbf{A1} & \textbf{A0} & \textbf{A1} & \textbf{A2} & \textbf{A0} & \textbf{A1} & \textbf{A2} \\
    \midrule
    \textbf{CP-SLAM} & 	\textbf{150} & 	\underline{150} & \underline{250} & \underline{250} & 	\underline{250} & 	\underline{250} & 	\underline{250} & 	\underline{250}  & 	\textbf{50} & 	\textbf{50} & - & 	\textbf{50} & 	\textbf{50}  & - \\
    \midrule
    \textbf{MAGiC-SLAM} & 300 & 300 & 500 & 500 & 500 & 500 & 500 & 500 & 100 & 100 & 100 & 100 & 100 & 100 \\
    \midrule
    \textbf{MAC-Ego3D} & \textbf{150} & \underline{150} & \underline{250} & 	\underline{250} & 	\underline{250} & 	\underline{250} & 	\underline{250} & 	\underline{250}	& 	\textbf{50} & 	\textbf{50} & \textbf{50} & \textbf{50} & \textbf{50} & \textbf{50} \\
    \midrule
    \textbf{Ours} & \underline{160} & \textbf{147} & \textbf{196} & \textbf{235} & \textbf{170} & \textbf{171} & \textbf{194} & \textbf{176} & \underline{84} & \underline{89} & \underline{94} & \underline{86} & \underline{89} & \underline{92} \\
    \bottomrule
    \end{tabular}%
    }
    \label{tab:keyframe_count}
\end{table}
\begin{table}
    \caption{\textbf{Total data transmitted in megabytes per agent across different scenes.} Comparison for Replica and Aria datasets. - indicates that CP-SLAM does not support more than two agents. \textbf{Bold} = best (least), \underline{underlined} = second best.}
    \centering
    \setlength\tabcolsep{4pt}
    \resizebox{\columnwidth}{!}{%
    \begin{tabular}{lccccccccccccccc}
    \toprule
    & \multicolumn{8}{c}{\textbf{Replica}} & \multicolumn{6}{c}{\textbf{Aria}} \\
    \cmidrule(lr){2-9} \cmidrule(lr){10-15}
    & \multicolumn{2}{c}{\textbf{Office-0}} & \multicolumn{2}{c}{\textbf{Apart-0}} & \multicolumn{2}{c}{\textbf{Apart-1}} & \multicolumn{2}{c}{\textbf{Apart-2}} & \multicolumn{3}{c}{\textbf{Room0}} & \multicolumn{3}{c}{\textbf{Room1}} \\
    \textbf{Methods} & \textbf{A0} & \textbf{A1} & \textbf{A0} & \textbf{A1} & \textbf{A0} & \textbf{A1} & \textbf{A0} & \textbf{A1} & \textbf{A0} & \textbf{A1} & \textbf{A2} & \textbf{A0} & \textbf{A1} & \textbf{A2} \\
    \midrule
    \textbf{CP-SLAM} & 376 & 376 & 626 & 626 & 627 & 627 & 626 & 626 & \textbf{45} & \textbf{44} & - & \textbf{43} & \textbf{43} & -\\
    \midrule
    \textbf{MAGiC-SLAM} & 2034 & 2070 & 3471 & 3474 & 3569 & 3543 & 3537 & 3549 & 538 & 531 & 515 & 541 & 530 & 512 \\
    \midrule
    \textbf{MAC-Ego3D} & 2061 & 2008 &  5027 & 4861 & 6551 & 3404 & 3448 & 3329 & 656 & 2305 & 2391 & 554  & 631 & 410 \\
    \midrule
    \textbf{Ours(Rendered depth)} & \textbf{98} & \textbf{108} & \textbf{187} & \underline{230} & \underline{247} & \underline{256} &\textbf{194} & \textbf{202} & {61} & {113} & \underline{79} & {89} & \underline{68} & \underline{56}  \\
    \midrule
    \textbf{Ours(Camera depth)} & \underline{170} & \underline{195} & \underline{193} & \underline{203} & \textbf{200} & \textbf{183} & \underline{237} & \underline{222} & \underline{55} & \underline{79} & \underline{65} & \underline{66} & {71} & \textbf{50} \\
    \bottomrule
    \end{tabular}%
    }
    \label{tab:submap_sizes}
\end{table}

\subsection{Ablation Study}

\Cref{tab:ablation_sizes} examines the individual contributions of different components for data transmitted in our system. We evaluate three data representation types: point clouds (following MAGiC-SLAM), rendered depth maps with pure 3D Gaussian representations, and camera depth maps with co-visible Gaussian removal from the first keyframe in the next submap. Additionally, we analyze the impact of keyframing and compaction strategies.

The ablation results demonstrate the complementary benefits of our proposed components. Keyframing provides substantial reductions in submap sizes across all data types, with the most significant improvements observed when transitioning from point cloud representations to 3D Gaussian-based approaches. The compaction strategy offers additional size reductions, though its impact varies depending on the underlying data representation. When examining the combination of techniques, rendered depth with both keyframing and compaction achieves the most compact representations across most scenarios.


\begin{table}
    \caption{\textbf{Total data transmitted in megabytes under different settings.} Transmitted data is reported w.r.t. the combination of settings: (1) Data type: Point cloud (\textbf{Pcd}, same as MAGiC-SLAM), rendered depth (\textbf{RD}, pure 3D Gaussian map) and camera depth (\textbf{CD}, 3D Gaussian map with depth image) (2) Keyframing (\textbf{K}) (3) Compaction (\textbf{C}). \textbf{Bold} = best (least), \underline{underlined} = second best.}
    \centering
    \setlength\tabcolsep{4pt}
    \resizebox{\columnwidth}{!}{%
    \begin{tabular}{lccccccccccccccc}
    \toprule
    & \multicolumn{8}{c}{\textbf{Replica}} & \multicolumn{6}{c}{\textbf{Aria}} \\
    \cmidrule(lr){2-9} \cmidrule(lr){10-15}
    & \multicolumn{2}{c}{\textbf{Office-0}} & \multicolumn{2}{c}{\textbf{Apart-0}} & \multicolumn{2}{c}{\textbf{Apart-1}} & \multicolumn{2}{c}{\textbf{Apart-2}} & \multicolumn{3}{c}{\textbf{Room0}} & \multicolumn{3}{c}{\textbf{Room1}} \\
    \textbf{Methods} & \textbf{A0} & \textbf{A1} & \textbf{A0} & \textbf{A1} & \textbf{A0} & \textbf{A1} & \textbf{A0} & \textbf{A1} & \textbf{A0} & \textbf{A1} & \textbf{A2} & \textbf{A0} & \textbf{A1} & \textbf{A2} \\
    \midrule
    \textbf{Pcd} & 2034 & 2070 & 3471 & 3474 & 3569 & 3543 & 3537 & 3549 & 538 & 531 & 515 & 541 & 530 & 512 \\
    \textbf{Pcd + C} & 2027 & 2057 & 3455 & 3452 & 3542 & 3522 & 3516 & 3533 & 525 & 516 & 505 & 529 & 520 & 504 \\
    \textbf{Pcd + K} & 1471 & 1760 & 1353 & 1373 & 1317 & 1221 & 1568 & 1441 & 256 & 386 & 354 & 325 & 269 & 278\\
    \textbf{Pcd + K + C} & 1466 & 1750 & 1330 & 1344 & 1290 & 1204 & 1545 & 1419 & 238 & 366 & 338 & 306 & 255 & 265 \\
    \midrule
    \textbf{RD} & 307 & 356 & 570 & 664 & 753 & 776 & 607 & 651 & 278 & 326 & 277 & 309 & 260 & 256 \\
    \textbf{RD + K} & 220 & 237 & 350 & 420 & 370 & 402 & 336 & 375 & 147 & 249 & 190 & 200 & 140 & 138 \\
    \textbf{RD + C} & \underline{137} & \underline{144} & 224 & 276 & 258 & 269 & \underline{230} & 254 & 160 & 194 & 159 & 182 & 149 & 144 \\
    \textbf{RD + K + C} & \textbf{98} & \textbf{108} & \textbf{187} & \underline{230} & 247 & 256 & \textbf{194} & \textbf{202} & \underline{61} & 113 & \underline{79} & \underline{89} & \textbf{68} & \underline{56} \\
    \midrule
    \textbf{CD} & 215 & 234 & 399 & 386 & 443 & 441 & 459 & 452 & 126 & 128 & 111 & 126 & 116 & 101 \\
    \textbf{CD + K} & 180 & 206 & 215 & 234 & \underline{227} & \underline{200} & 259 & 242 & 74 & \underline{98} & 81  & 85 & 86 & 63 \\
    \textbf{CD + C} & 208 & 220 & 384 & 365 & 418 & 422 & 438 & 436 & 113 & 113 & 101 & 113 & 106 & 93 \\
    \textbf{CD + K + C} & 170 & 195 & \underline{193} & \textbf{203} & \textbf{200} & \textbf{183} & 237 & \underline{222} & \textbf{55} & \textbf{79} & \textbf{65} & \textbf{66} & \underline{71} & \textbf{50}\\
    \bottomrule
    \end{tabular}%
    }
    \label{tab:ablation_sizes}
\end{table}

We would also like to mention that FPFH + RANSAC coarse registration is essential for map merging when the initial relative poses between agents are unknown. Because ICP can only provide fine adjustment of relative poses between submaps, without reasonable initial guesses, it will output incorrect loop-edge pose estimates, leading to map merging failures during PGO, as shown in~\cref{tab:rendering_results} and~\cref{fig:topview_comparison}.

\section{Conclusion}
We present Coko-SLAM, a multi-agent 3DGS SLAM system. Our method achieves map merging without initial relative poses between agents while maintaining rendering quality comparable to or better than existing approaches, and reduces transmitted data by up to 95\% through strategic keyframing and compaction. A current limitation is the reliance on rendered depth images for inter-robot registration, which can degrade at lower camera resolutions, as observed on the Aria dataset.

Despite these encouraging results, there is further space for improvement. Future work will focus on developing fully decentralized architectures to eliminate server dependencies and investigating faster tracking methods to further improve real-time performance in resource-constrained environments. We would also like to explore the possibility of registering 3D Gaussian submaps without rendering for better map merging performance.

\section*{Acknowledgements}
This work was supported by the Natural Sciences and Engineering Research Council of Canada (NSERC), Funding Reference Number: CGV192714.

\bibliographystyle{unsrt}  

\bibliography{ref}

\end{document}